\documentclass{article}
\usepackage{spconf,amsmath,amssymb,graphicx,booktabs,tabularx,hyperref}
\usepackage{stfloats}

\title{Same Words, Different Actions: Paired Turn-Taking Evaluation under Rewritten Dialogue Contexts}
\name{\shortstack[c]{%
Shuofeng Zhao$^{\star\dagger}$, Hongwei Cai$^{\star}$, Wenke Fan, Qingxiang Guo, Zhou Wang, Dawei Yang, \\
Zhiyang Zhou, Yingxin Shang, Weixu Wang, Lin Yang, Shuran Zhou, Yang Song%
\thanks{$^{\star}$ Equal contribution.}%
\thanks{$^{\dagger}$ Corresponding author. E-mail: zhaoshuofeng@zuoyebang.com}%
}}
\address{Zuoyebang Education Technology}

\begin{document}
\ninept
\maketitle
\begin{abstract}
    Real-time spoken dialogue systems must distinguish interruptions that
    require yielding the floor from backchannels that permit continued speaking.
    Existing benchmarks typically score events independently and may
    therefore assign high scores to systems with fixed action preferences
    rather than context-sensitive decision policies. We
    introduce ECHO, a paired diagnostic benchmark for Chinese turn-taking 
    evaluation. ECHO pairs examples with the same overlap transcript but
    contrasting preceding multi-turn dialogue contexts, with one requiring
    \textsc{Yield} and the other \textsc{Keep}. It additionally includes off-talk
    examples for diagnosing unnecessary yielding. We introduce pair accuracy,
    which requires correct decisions on both members of a pair and assigns no
    credit to constant-action policies. Experiments on four speech systems 
    show that three exhibit a severe over-yielding bias: they correctly keep 
    the floor on fewer than 13\% of backchannels, resulting in near-zero pairwise 
    success rates ($\le$4\%). While the remaining system remains comparatively 
    balanced across contexts, these findings broadly demonstrate that 
    interruption-only evaluation can severely overestimate practical turn-taking 
    reliability.
\end{abstract}
\begin{keywords}
  full-duplex spoken dialogue, turn-taking, interruption detection,
  overlapping speech, diagnostic evaluation
\end{keywords}
\section{Introduction}

Real-time spoken dialogue systems must decide how to respond when a user
speaks while the system is still producing speech. This capability is
required not only by end-to-end full-duplex models, but also by modular
systems that support barge-in through a semantic VAD, turn-state
predictor, or interruption detector. Regardless of implementation, speech
activity detection alone is insufficient: overlapping user speech may be
a correction or request that claims the conversational floor, but it may
also be an acknowledgment, an affiliative comment, or speech addressed to
the user themselves or a third party. Yielding to every overlap produces
fragmented interactions, whereas ignoring genuine floor claims prevents
users from correcting or redirecting the system. A central overlap-handling
decision is therefore whether the system should stop speaking
(\textsc{Yield}) or continue its current turn (\textsc{Keep}).

We study this decision as a common behavioral component of real-time
spoken dialogue systems, independently of whether it is implemented by an
external semantic turn detector or by the internal control mechanism of a
full-duplex model. Specifically, given a predefined multi-turn dialogue
context and an overlapping user utterance, we evaluate whether a system
produces the appropriate \textsc{Yield}/\textsc{Keep} action. We refer to
this task as \emph{context-conditioned overlap handling}. Some evaluated
systems process the audio through their native streaming interfaces, but
our protocol replays fixed dialogue trajectories. ECHO therefore
diagnoses event-level overlap decisions rather than closed-loop
interaction quality, user adaptation, or response latency.

Recent benchmarks have expanded the evaluation of overlap handling in
spoken dialogue systems. Full-Duplex-Bench and its extensions cover
interruptions, backchannels, pauses, non-target speech, and multi-turn
interaction
quality~\cite{Lin2025FullDuplexBenchAB,Lin2025FullDuplexBenchVE,lin2026fullduplexbenchv2multiturnevaluationframework};
the ICASSP 2026 HumDial Challenge evaluates responses to genuine
interruptions and non-interruptive
feedback~\cite{wang2026fullduplexinteractionspokendialogue}; and TurnBench examines false
interruptions across conversation
types~\cite{jiang2026turnbenchmultidomainbenchmarkturntaking}. Related
work studies robustness to third-party speech~\cite{lee2026still},
semantic-aware interruption detection~\cite{xia2026semanticawareinterruptiondetectionspoken}, semantic
VAD, streaming state prediction, and context-aware turn
management~\cite{Wang2026FastTurnUA,Wu2025PhoenixVADSS,yan2026soulxduplugplugandplaystreamingstate,Zhang2025LLMEnhancedDM}.
These decision mechanisms are also integrated into end-to-end full-duplex
architectures~\cite{yu2025salmonnomni,liu2026hierarchicalacousticsemanticmodelingmodality,Cui2026MiniCPMo4T}.

Existing resources cover a broad range of interaction phenomena, but
their evaluation events are generally collected from natural
conversations or generated independently from scenario scripts. The
overlapping utterance and its preceding dialogue context therefore
typically vary together. As a result, a correct prediction does not show
whether a system would change its action if the same insertion occurred
under a different context. Moreover, some non-interruptive subsets are
dominated by conventional feedback expressions. In the analyzed Chinese
split of SID-Bench~\cite{xia2026semanticawareinterruptiondetectionspoken}, 81.8\% of non-interruptive
instances consist entirely of its fifteen most frequent characters; in
the analyzed Easy Turn~\cite{li2025easyturnintegratingacoustic} test set,
the backchannel and turn-taking subsets have disjoint vocabularies
(Table~\ref{tab:lexical}). Such distributions may allow systems to
exploit insertion-level lexical regularities without demonstrating
context-sensitive interpretation.
\begin{table}[b!]
    \centering
    \caption{Lexical closure of non-interruptive feedback. }
    \label{tab:lexical}
    \footnotesize
    \begin{tabular}{lrrr}
        \toprule
        Set & Inst. & Char.\ voc. & Top-15-char only \\
        \midrule
        SID-Bench (zh) & 494 & 124 & 81.8 \\
        Easy Turn      & 100 &  67 & 26.0 \\
        ECHO (ours)    & 183 & 400 &  0.0 \\
        \bottomrule
    \end{tabular}
\end{table}

\begin{figure}[t]
    \centering
    \includegraphics[width=\columnwidth]{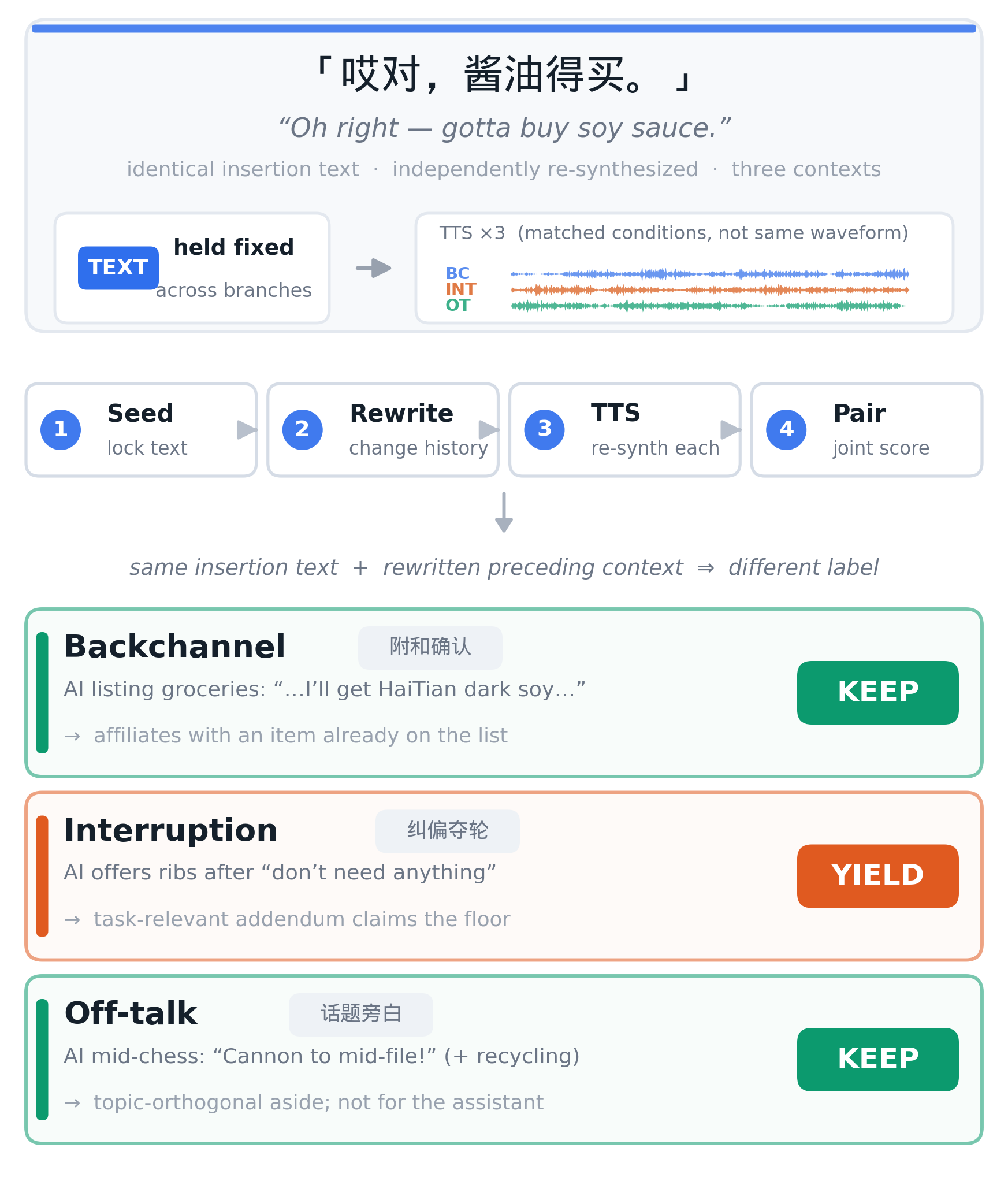}
    \caption{One insertion, three required actions. All three ECHO branches
      share the same user insertion transcript, glossed ``Oh right - gotta buy soy sauce''. In each branch, the insertion is placed at the
      annotated overlap position located by forced alignment, while the
      preceding multi-turn dialogue context is rewritten. The rewrite
      changes the interactional role from backchannel to interruption to
      off-talk, and with it the target action from \textsc{Keep} to
      \textsc{Yield} to \textsc{Keep}. Lexical form is therefore
      uninformative, and a system must decide before the assistant track
      terminates.}
    \label{fig:motivation}
\end{figure}

More substantive overlapping utterances can require different actions
depending on the preceding dialogue. For example, ``Oh right---gotta
buy soy sauce'' can be a non-floor-claiming affiliation when the
assistant is already listing groceries and mentioning dark soy sauce,
but a floor-claiming addendum when the user has just declined further
purchases and the assistant nonetheless offers to buy ribs. It may
also be speech addressed to the user themselves or to a third party
while the assistant is mid-chess and discussing recycling
(Fig.~\ref{fig:motivation}). The insertion transcript alone therefore
does not determine whether the system should yield or continue. This
raises our central question: can a spoken dialogue system produce
appropriate \textsc{Yield}/\textsc{Keep} decisions across rewritten
contexts that share the same insertion transcript?

We introduce ECHO (Evaluating Context-conditioned Handling of Overlaps), a
paired diagnostic set of Chinese multi-turn dialogues. Linked ECHO
instances share the same insertion transcript but use rewritten preceding
contexts that assign the insertion different interaction roles:
interruption, backchannel, or off-talk. These roles correspond to
\textsc{Yield}, \textsc{Keep}, and \textsc{Keep}, respectively. This
transcript-matched construction removes insertion-text differences within
each pair, although the independently synthesized waveforms are not fully
acoustically matched. We further propose Pairwise Action Success Rate
(PASR), which requires jointly correct actions on an action-flip pair and
assigns zero credit to deterministic constant-action policies. Evaluating
four speech systems through their supported interfaces, we find that
three yield correctly on many interruption instances but keep the floor
on fewer than 13\% of backchannels, resulting in
interruption--backchannel PASR values of at most 4\%. These results show
that interruption accuracy alone is insufficient to characterize
balanced overlap handling.

\section{ECHO Dataset Construction}
\label{sec:echo}

ECHO follows an insertion-text-matched contrastive design inspired by
minimal-pair testing, a strategy also used to probe whether audio language
models genuinely attend to the acoustic evidence they are given
\cite{Xiong2026DEAFAB}.

\subsection{Interaction Roles and System Actions}
\label{sec:taxonomy}

ECHO defines three interaction roles by intended addressee and target
floor action, as summarized in Table~\ref{tab:interaction_roles}.

\begin{table}[t]
\centering
\small
\setlength{\tabcolsep}{2.8pt}
\renewcommand{\arraystretch}{1.1}
\begin{tabular}{@{}lcccc@{}}
\toprule
& \multicolumn{2}{c}{\textbf{Addressee}}
& \textbf{Claims} & \textbf{Target} \\
\cmidrule(lr){2-3}
\textbf{Role}
& \textbf{Asst.}
& \textbf{Self/other}
& \textbf{floor}
& \textbf{action} \\
\midrule
Interruption & \checkmark & --         & \checkmark & \textsc{Yield} \\
Backchannel  & \checkmark & --         & --         & \textsc{Keep}  \\
Off-talk     & --         & \checkmark & --         & \textsc{Keep}  \\
\bottomrule
\end{tabular}
\caption{Interaction roles by addressee, floor intent, and target action.}
\label{tab:interaction_roles}
\end{table}

An \textbf{interruption} is directed to the assistant and introduces a
request, correction, task obstacle, condition change, or control
instruction that requires immediate handling; its target action is
therefore \textsc{Yield}. A \textbf{backchannel} is also
assistant-directed but does not claim the floor. We use the term broadly
for acknowledgments, affective feedback, affiliative responses, and
brief collaborative completions, with \textsc{Keep} as the target
action. In contrast, \textbf{off-talk} is directed to the speaker
themself or a third party and is likewise assigned \textsc{Keep}. Stage
directions used during off-talk generation are never exposed to the
evaluated models.

We retain the original three-way labels alongside the binary action
labels because the two \textsc{Keep} roles differ downstream:
backchannels are assistant-directed feedback, whereas off-talk should
generally not be committed to the dialogue state as a system-directed
user turn.

\subsection{Context-Rewritten Contrast Generation}
\label{sec:context_generation}

We first randomly sample Chinese multi-turn dialogue skeletons from the
synthesized dialogue data constructed in
DuplexDrama~\cite{guo2026duplexdramasynthesizeddialoguedataset}, which
covers diverse interlocutor relationships, locations, everyday tasks, and
conversational situations. DeepSeek-V4-Pro~\cite{deepseekai2026deepseekv4} then selects
candidate overlap positions and generates initial backchannel or off-talk
insertions according to the dialogue state and character roles. We retain
only insertions that can plausibly receive an alternative interactional
interpretation under a rewritten context; we do not force every insertion
to realize all three roles.

Given a selected insertion, Claude-3.5-Sonnet~\cite{anthropic2024claude35sonnet}
rewrites the dialogue history up to and including the assistant utterance
being overlapped while keeping the insertion text unchanged, under a
prompt that enforces the role definitions of Sec.~\ref{sec:taxonomy}.
Let $u_i$ denote the shared insertion across two context variants $c_i^{(a)}$ and $c_i^{(b)}$. For this contrast pair, we enforce $u_i^{(a)}=u_i^{(b)}=u_i$ and $y_i^{(a)}\neq y_i^{(b)}$, and each linked
instance is represented as $(c_i^{(k)},u_i,y_i^{(k)})$. A group may
contain two or all three role variants. The equality constraint applies to
the insertion text, not to its synthesized waveform.

We first use DeepSeek-V4-Pro to screen candidate instances for consistency among the rewritten history, the insertion, and the intended interactional role. The screened candidates are then manually reviewed by human annotators according to the same criteria. Overall, 97.45\% of the reviewed candidates are judged to be consistent with their target labels.

\subsection{Speech Synthesis and Overlap Rendering}
\label{sec:audio_rendering}

All dialogue turns are independently synthesized with IndexTTS2
  \cite{zhou2025indextts2breakthroughemotionallyexpressive} and rendered as
  speaker-separated dual-channel audio. We apply the same synthesis and
  post-processing pipeline to all examples, including speaker-prompt RMS
  normalization, bounded speaking-rate normalization, forced-alignment-based
  overlap placement, and role-dependent overlap rendering.

  Within each linked group, the insertion text, emotion condition, and TTS
  inference configuration are held fixed, while the preceding multi-turn
  context and the intended interaction role are changed. The role determines
  the rendered interaction: backchannel and off-talk speech is overlaid
  without modifying the assistant track, whereas an interruption receives
  onset emphasis and causes the assistant track to fade to silence after a
  short reaction interval. Speaker-reference audio may differ across paired
  instances, and all utterances are synthesized independently; paired
  insertions are therefore controlled in lexical and selected synthesis
  conditions, but are not waveform-identical or fully acoustically matched.
  The faded assistant waveform is excluded from evaluated model inputs, so
  post-decision floor release cannot serve as a label cue.

\subsection{Dataset Organization and Scope}
\label{sec:dataset_statistics}


ECHO contains 266 groups, 549 unique audio instances, and 300 pair
relations, distributed over two-role and three-role groups as reported in
Table~\ref{tab:data_statistics}. The three roles are balanced at the
instance level (183 each), and pair relations are balanced by construction
with 100 pairs for each of the interruption--backchannel,
interruption--off-talk, and backchannel--off-talk contrasts. We release the
audio, model-observable dialogue text, original scenario labels, binary
action labels, event timestamps, and group/pair metadata at
\url{https://huggingface.co/datasets/shuofeng123/ECHO}.

\begin{table}[t]
    \centering
    \caption{ECHO dataset statistics. Groups, unique samples, and pair
    relations are counted separately so that three-role groups are not
    counted multiple times at the instance level. Each of the 17
    three-role groups induces all three pair contrasts.}
    \label{tab:data_statistics}
    \resizebox{\columnwidth}{!}{
    \begin{tabular}{lrrr}
        \toprule
        Group type & Groups & Unique samples & Pair relations \\
        \midrule
        Interruption--Backchannel only & 83 & 166 & 83 \\
        Interruption--Off-talk only     & 83 & 166 & 83 \\
        Backchannel--Off-talk only      & 83 & 166 & 83 \\
        Three-role groups               & 17 &  51 & 51 \\
        \midrule
        Total                           & 266 & 549 & 300 \\
        \bottomrule
    \end{tabular}}
\end{table}

Non-interruptive feedback in prior test sets is separable by lexical form
alone, and ECHO removes that shortcut by construction. All 100
interruption--backchannel pairs carry identical insertion text, each of
the 183 backchannel instances is lexically unique, none is built only from
the fifteen most frequent characters, and the three roles have matched
length distributions (6.3, 6.2, and 6.6 characters on average), so the
insertion text carries no discriminative information within a pair
(Table~\ref{tab:lexical}).

\section{Experiments}
\label{sec:experiments}

\subsection{Evaluated Systems and Native Interfaces}
\label{sec:models}

\begin{table}[!tp]
    \centering
    \caption{Native model interfaces and event-level action extraction.
    Context availability differs across systems, and assistant-side
    context is provided only when supported by the native interface.}
    \label{tab:model_interfaces}
    \resizebox{\columnwidth}{!}{
    \begin{tabular}{llll}
        \toprule
        Model & User-side input & Assistant-side input & Action extraction \\
        \midrule
        Easy Turn
        & Target insertion
        & None
        & Native state mapping \\

        SoulX-Duplug
        & Last 2 user turns + insertion
        & None
        & Trigger within event window \\

        Lychee-FD
        & Up to 5 user turns + insertion
        & History + planned utterance
        & Stop event during insertion \\

        MiniCPM-o 4.5
        & Full history + insertion
        & Teacher-forced planned utterance
        & Stop event during insertion \\

        Gemini
        & Insertion transcript
        & History + spoken prefix
        & Prompted role prediction \\
        \bottomrule
    \end{tabular}}
\end{table}

We evaluate four speech systems and one text-only language model
(Table~\ref{tab:model_interfaces}). Each system is tested through its
supported interface, resulting in different context availability,
modalities, and native decision mechanisms. The results therefore
constitute a behavioral evaluation under native interfaces rather than a
controlled comparison of model architectures or context representations.

\noindent\textbf{Easy Turn}~\cite{li2025easyturnintegratingacoustic}
is a modular turn-state predictor with four native outputs. We map
\texttt{complete} and \texttt{incomplete} to \textsc{Yield}, and
\texttt{backchannel} and \texttt{offtalk} to \textsc{Keep}, because the
former two indicate a user floor claim. As it receives only the target
insertion, Easy Turn serves as a context-free local baseline.

\noindent\textbf{SoulX-Duplug}%
~\cite{yan2026soulxduplugplugandplaystreamingstate}
is a streaming state predictor operating on 160-ms audio chunks. We
record \textsc{Yield} if its interruption state is triggered during the
insertion or within 1\,s after the insertion ends, and \textsc{Keep}
otherwise.

\noindent\textbf{Lychee-FD}%
~\cite{liu2026hierarchicalacousticsemanticmodelingmodality}
is a full-duplex dialogue model. 
We monitored the model’s speaking/listening state within its native 400-ms decision window. Interruptions were correct if speech stopped, whereas backchannels and off-talk were correct if speech continued.

\noindent\textbf{MiniCPM-o 4.5}~\cite{Cui2026MiniCPMo4T}
is an end-to-end multimodal dialogue model. We force-align the final
assistant utterance and teacher-force its text while streaming the user
audio up to the target event. As with Lychee-FD, a stop event during the
insertion is mapped to \textsc{Yield}; continued generation is mapped to
\textsc{Keep}.

\noindent\textbf{Gemini-3.1-Pro-Preview}%
~\cite{google2026gemini31pro}
predicts one of the three ECHO roles from text. It receives the observable
dialogue history, the spoken prefix of the current assistant utterance,
and the insertion transcript, but no audio, stage directions, unspoken
assistant content, or post-event reference responses. It is included as
a semantic reference rather than a modality-matched baseline or an upper
bound.

\subsection{Action Alignment and Metrics}
\label{sec:action_alignment}

\begin{table*}[!t]
    \centering
    \caption{Binary system-action evaluation on ECHO. Input interfaces are
    given in Table~\ref{tab:model_interfaces}. All sample-level
    accuracies are computed over unique instances. Gemini's three-way
    predictions are mapped to \textsc{Yield}/\textsc{Keep} before
    computing this table. }
    \label{tab:action_results}
    \resizebox{\textwidth}{!}{
    \begin{tabular}{lrrrrrrrr}
        \toprule
        Model
        & Int. \textsc{Yield}
        & BC \textsc{Keep}
        & OT \textsc{Keep}$^\dagger$
        & Macro
        & Overall
        & PASR$_{I-B}$
        & PASR$_{I-O}$
        & PKC$_{B-O}$ \\
        \midrule
        Easy Turn
        & \textbf{98.91}
        & 1.09
        & 0.55
        & 33.52
        & 33.52
        & 0.00 & 0.00 & 1.00 \\

        SoulX-Duplug
        & 91.26
        & 6.56
        & 15.30
        & 37.71
        & 37.71
        & 4.00 & 12.00 & 2.00 \\

        Lychee-FD
        & 55.19
        & 12.02
        & 8.20
        & 25.14
        & 25.14
        & 4.00 & 6.00 & 0.00 \\

        MiniCPM-o 4.5
        & 63.39
        & 65.57
        & 53.55
        & 60.84
        & 60.84
        & 54.00 & \textbf{53.00} & 49.00 \\

        Gemini reference
        & 90.71
        & \textbf{86.34}
        & \textbf{71.58}
        & \textbf{82.88}
        & \textbf{82.88}
        & \textbf{79.00} & 52.00 & \textbf{70.00} \\
        \midrule
        \multicolumn{9}{l}{\textit{Three-way role prediction (same
        reference, original role labels; PRSR in place of PASR/PKC)}} \\
        Gemini reference
        & 90.71
        & 85.25
        & 71.04
        & 82.33
        & 82.33
        & \multicolumn{3}{c}{66.00 (PRSR)} \\
        \bottomrule
    \end{tabular}}
    \vspace{1mm}

    \begin{minipage}{0.96\textwidth}
    \footnotesize
    $^\dagger$Keep rate on scenario-labeled off-talk, interpreted as a
    scenario-based action diagnostic rather than an unambiguous
    false-trigger estimate.
    \end{minipage}
\end{table*}

We map interruption to \textsc{Yield} and backchannel and off-talk to
\textsc{Keep} (Sec.~\ref{sec:taxonomy}), over all unique instances. For speech
models without an explicit off-talk state, an off-talk instance counts as
correct whenever the system continues its current turn, so no
fine-grained off-talk recognition is required. Gemini's three-way
predictions are mapped to the same binary space before action metrics are
computed; predicting off-talk for a backchannel is therefore wrong in the
three-way analysis but right in the binary one. We additionally evaluate
Gemini in the original three-way role space to test whether the intended
roles can be recovered from explicit textual context.

We report action accuracy over unique samples, separately for each
original label: $\mathrm{Acc}_{I}$ is the correct \textsc{Yield} rate on
interruptions, and $\mathrm{Acc}_{B}$ and $\mathrm{Acc}_{O}$ the correct
\textsc{Keep} rates on backchannel and off-talk. Macro accuracy averages
the three, and each unique sample is counted once even when it belongs to
a three-role group.

Sample-level accuracy can overestimate reliability when a model strongly
favors a single action. We therefore define the \textbf{Pairwise Action
Success Rate} (PASR) over action-flip pairs. For
$X\in\{I\text{-}B,I\text{-}O\}$, let $\mathcal{P}_X$ denote the set of
pairs of type $X$. The type-specific PASR is
\begin{equation}
\mathrm{PASR}_X
=
\frac{1}{|\mathcal{P}_X|}
\sum_{(p,q)\in\mathcal{P}_X}
\mathbb{1}
\left[
\hat a_p=a_p
\land
\hat a_q=a_q
\right],
\label{eq:pasr}
\end{equation}
where $a_s\in\{\textsc{Yield},\textsc{Keep}\}$ and $\hat a_s$ denote
the target and predicted actions for instance $s$, respectively. Each
pair in $\mathcal{P}_X$ contains one interruption instance and one
backchannel or off-talk instance. Thus, a pair is counted as successful
only if the model predicts \textsc{Yield} for the interruption and
\textsc{Keep} for its paired counterpart. Consequently, deterministic
constant-\textsc{Yield} and constant-\textsc{Keep} policies both obtain
zero PASR.

For backchannel--off-talk pairs, whose target actions are both
\textsc{Keep}, we report \textbf{Pairwise Keep Consistency} (PKC), the
proportion of pairs for which the model predicts \textsc{Keep} for both
instances. PASR evaluates appropriate action switching across
action-flip pairs, whereas PKC evaluates consistent floor maintenance
across two non-floor-claiming roles. A constant-\textsc{Keep} policy
maximizes PKC but obtains zero PASR, while a constant-\textsc{Yield}
policy obtains zero on both metrics. The two metrics should therefore be
interpreted jointly.

For three-way role prediction, we additionally report the
\textbf{Pairwise Role Success Rate} (PRSR) over all three pair types. A
pair contributes to PRSR only when both instances are assigned their
correct original role labels.

\subsection{Results and Analysis}
\label{sec:action_results}

Table~\ref{tab:action_results} summarizes the sample-level action
accuracies and pair-level results for all evaluated systems.

\noindent\textbf{Finding 1: same-transcript action flips remain difficult
for most evaluated speech systems.}
On interruption--backchannel pairs, whose members share the same
insertion transcript but require opposite actions, Easy Turn,
SoulX-Duplug, and Lychee-FD achieve PASR values of 0.00\%, 4.00\%, and
4.00\%, respectively. MiniCPM-o 4.5 performs substantially better at
54.00\%, while the text-only Gemini reference reaches 79.00\%. These
results show that correct handling of one member of a matched contrast
does not generally imply success on its action-flipped counterpart.
The Gemini result provides additional evidence that many intended action
contrasts are recoverable from the observable textual context.

\noindent\textbf{Finding 2: low paired success is associated with a
pronounced \textsc{Yield} preference.}
Easy Turn and SoulX-Duplug correctly yield on 98.91\% and 91.26\% of
interruptions, respectively, but keep the floor on only 1.09\% and
6.56\% of backchannels. Lychee-FD likewise keeps the floor on only
12.02\% of backchannels, although its interruption accuracy is lower
at 55.19\%. Thus, interruption-only accuracy would portray Easy Turn
and SoulX-Duplug favorably despite their near-zero success on matched
interruption--backchannel contrasts. MiniCPM-o 4.5 is more balanced,
with interruption and backchannel accuracies of 63.39\% and 65.57\%,
respectively. PASR therefore complements class-conditioned accuracy by
requiring both members of an action-flip pair to be correct and assigning
no credit to deterministic constant-action policies.

\noindent\textbf{Interpretation caveat.}
PASR measures joint behavioral success on transcript-matched pairs, but
does not by itself establish that action differentiation is caused
exclusively by the rewritten dialogue context. Paired insertions are
independently synthesized, and the onset emphasis described in
Sec.~\ref{sec:audio_rendering} is applied only to interruptions. This cue
cannot directly explain the low \textsc{Keep} rates on untreated
backchannels and off-talk, but it may facilitate interruption detection
and contribute to PASR. A waveform-matched, gain-controlled evaluation
would be required to isolate contextual sensitivity.

\section{Conclusion}

We introduced ECHO, a paired diagnostic benchmark for overlap handling
in Chinese spoken dialogue systems. ECHO holds the insertion transcript
fixed across rewritten dialogue contexts and evaluates joint action
correctness on matched pairs. Results on four speech systems show that
interruption accuracy alone is insufficient to characterize balanced
overlap handling: three systems keep the floor on fewer than 13\% of
backchannels and achieve at most 4\% PASR on
interruption--backchannel pairs, although two of them exceed 90\%
interruption accuracy. These findings motivate reporting
class-conditioned \textsc{Keep} rates and pair-level success alongside
interruption accuracy. 

\noindent\textbf{Limitations.}
ECHO is a synthetic, event-level diagnostic rather than an evaluation of
natural or closed-loop dialogue. Paired instances match the insertion
transcript but not the waveform, and interruptions contain role-specific
onset emphasis; PASR therefore measures behavioral success on the
released stimuli but cannot isolate the effect of dialogue context from
correlated acoustic cues. Moreover, replaying fixed dialogue trajectories,
including teacher-forced assistant turns where required, cannot simulate
real-time full-duplex interaction or evaluate latency, mutual adaptation,
post-yield recovery, and subsequent response quality. 

\section{Acknowledgments}

Generative AI is used in this work in two capacities, both disclosed in
accordance with IEEE policy. As construction tools, the models named in
Sec.~\ref{sec:echo} generate the ECHO material: dialogue skeletons,
context rewrites, and synthesized audio. In manuscript
preparation, the authors used LLM for language
polishing and assisted drafting. No AI system contributed to the
experimental design, the reported results, or the scientific claims, and
the authors take full responsibility for the content of this publication.

\bibliographystyle{IEEEbib}
\bibliography{strings,refs}

\end{document}